\documentclass{article}
\PassOptionsToPackage{numbers, compress}{natbib}

\usepackage[preprint]{neurips_2026}

\usepackage[utf8]{inputenc} 
\usepackage[T1]{fontenc}    
\usepackage{hyperref}       
\usepackage{url}            
\usepackage{booktabs}      
\usepackage{amsfonts}       
\usepackage{nicefrac}       
\usepackage{microtype}     
\usepackage{xcolor}         

\usepackage{amsmath}
\usepackage{graphicx}

\newcommand{\method}{PhyGenHOI}
\definecolor{deepblue}{RGB}{24, 0, 173}      
\definecolor{orange}{RGB}{255,127,80}  
\definecolor{green}{RGB}{0,128,0}      
\definecolor{royalblue}{RGB}{65, 105, 225}
\usepackage{wrapfig}
\usepackage{amssymb}

\title{\method{}: Physically-Aware 4D Generation of Dynamic Human-Object Interactions}

\author{%
  Omer Benishu \quad Gal Fiebelman \quad Sagie Benaim \\[0.2cm]
  Hebrew University of Jerusalem \\
  \vspace{0.07cm} 
}

\begin{document}

\maketitle

\begin{abstract}
We address the task of generating physically accurate and visually faithful 4D Human-Object Interaction (HOI). Given a static 3D human and target object represented as 3D Gaussian Splats (3DGS), our goal is to synthesize dynamic scenes where the human actively engages with the object through actions, such as punching or kicking, in accordance with a given input text. 
To this end, we introduce \method{}, a novel framework that couples generative human motion with an explicit physical object simulation. 
We model the human as a \textit{semantic agent} driven by a Motion Diffusion Model (MDM) and the object as a \textit{physical agent} simulated via the Material Point Method (MPM), utilizing 3D Gaussians as a unified, differentiable representation. 
We supervise their interaction through three coupled mechanisms: (1) A \textit{Windowed Attraction Loss} that temporally synchronizes generative motion to intercept the object; (2) A \textit{Contact-Driven Re-simulation} step that triggers physically consistent momentum transfer upon impact; and (3) A \textit{Masked Video-SDS} objective that injects video-based priors to enhance contact fidelity. 
Experiments show \method{} generates physically consistent 4D HOI across diverse actions, humans, and objects, outperforming baselines. Project page and videos: \url{https://omerbenishu.github.io/PhyGenHOI/}

\end{abstract}

\begin{figure}[h]
    \centering
    \input{figures/teaser} 
\end{figure}

\section{Introduction}
\label{sec:intro}

Synthesizing dynamic human-object interactions that are both visually faithful and physically plausible is a fundamental challenge in computer graphics, with critical applications in animation, gaming, and immersive virtual reality.
To this end, we consider the task of generating physically accurate and visually faithful 4D Human-Object Interaction (HOI). 
Specifically, given a static 3D human and a static target object, both represented as 3D Gaussian Splats (3DGS)~\cite{kerbl20233d}, our goal is to synthesize a dynamic 4D scene where the human actively engages with a dynamic object, such as kicking a soccer ball or pushing a file cabinet, in accordance with an input text. We aim to produce human and object motion that is both visually faithful and physically plausible, capturing the causal interplay of forces and collisions. By leveraging the explicit 3D Gaussians, we ensure that the resulting 4D content not only respects the laws of physics but also supports efficient rendering from novel viewpoints.

Despite the rapid evolution of text-to-4D generation approaches \cite{bahmani20244d, zhao2023animate124, ren2023dreamgaussian4d}, a critical dichotomy persists between semantic coherence and physical fidelity. 
On one hand, \textit{purely generative approaches} such as
4DFY~\cite{bahmani20244d} distill motion directly from large-scale video priors. 
While these methods excel at synthesizing diverse open-world scenarios, they fundamentally lack an underlying model of physics, frequently producing causal anomalies like ``ghosting'' artifacts where objects react before contact. 
On the other hand, \textit{kinematic frameworks} like AvatarGO~\cite{cao2025avatargo} and InterDreamer~\cite{xu2024interdreamer} introduce structured human priors (e.g., SMPL~\cite{SMPL:2015}) to ensure anatomical consistency. However, these methods typically reduce interaction to a geometric constraint, treating the target object as a ``static prop'' or a rigid accessory, failing to capture dynamic forces like ballistic momentum transfer. Similarly, recent 3D asset animation methods~\cite{wu2025animateanymesh, sun2025animus3d} animate individual entities but lack the coupled interaction logic required for human-object contact.

To bridge this gap, we introduce \method{}, generating 4D human-object interactions that are both semantically responsive and physically grounded. We devise a unified framework where 3D Gaussian Splatting serves as the common substrate for coupling semantic generation with physical simulation. To ensure kinematic fidelity, we model the human as an active agent driven by an SMPL-constrained Motion Diffusion Model (MDM), which provides a robust semantic prior for generating diverse, text-aligned actions. Conversely, we treat the object as a reactive physical agent by mapping its Gaussian kernels directly to particles in a differentiable Material Point Method (MPM) simulator, enforcing physically consistent object trajectories and deformations.

To coordinate these distinct agents into a cohesive interaction, we leverage three targeted mechanisms. 
First, to synchronize the human's semantic intent with the object's position, we propose a \textit{Windowed Attraction Loss} that spatially and temporally guides the generative motion to intercept the target. 
Second, to ensure physical causality, we implement \textit{Contact Detection and MPM Re-simulation}; upon detecting collision, the object's trajectory is explicitly updated to reflect realistic momentum transfer and material deformation. 
Finally, we apply a \textit{Temporally-Masked Video-SDS} that injects rich visual priors specifically around the contact frames, enhancing interaction fidelity without disrupting the physically grounded motion. Our framework targets actions involving discrete momentum transfer upon contact, such as kicking, punching, and pushing.

We validate our framework against state-of-the-art generative (4DFY~\cite{bahmani20244d}) and animation (AnimateAnyMesh~\cite{wu2025animateanymesh}) baselines across a suite of dynamic interaction scenarios. Our method eliminates the ghosting and interpenetration artifacts of purely generative models while producing dynamic object responses that animation methods cannot capture, achieving superior performance in text alignment, physical plausibility, contact quality, and visual fidelity.

\begin{figure*}

    \centering
    \includegraphics[width=\textwidth]{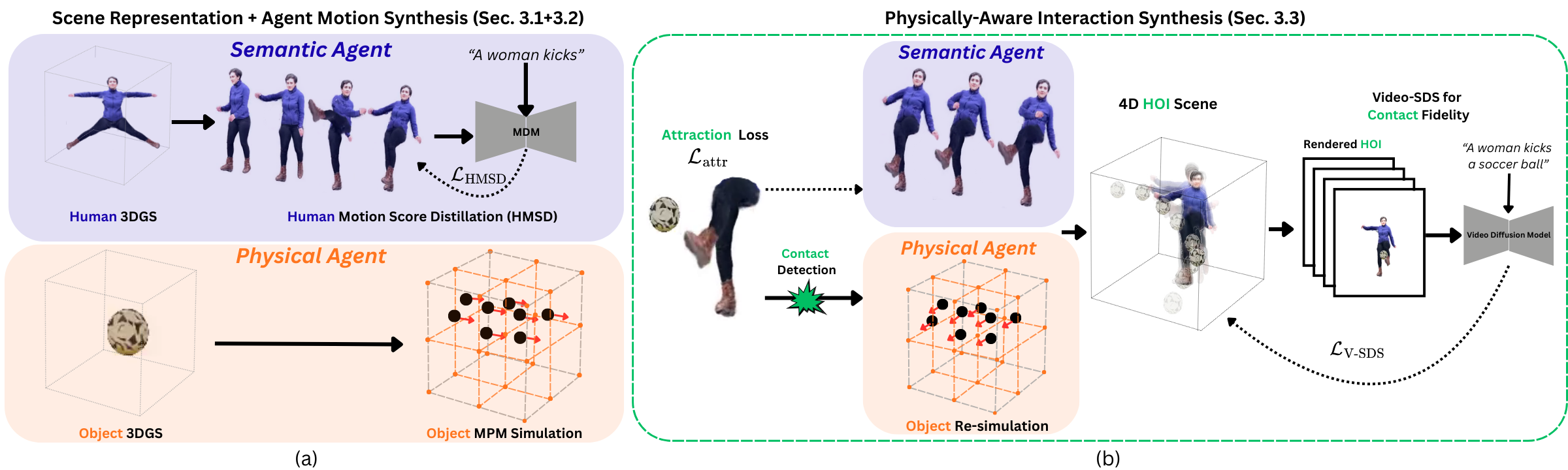} 

    \caption{\textbf{Overview of \method{}.} \textbf{(a) Scene Representation + Agent Motion Synthesis (Sec.~\ref{sec:representation}+\ref{sec:motion}):} Given a 3DGS \textcolor{deepblue}{human} and 3DGS \textcolor{orange}{object}, we treat the human as a \textcolor{deepblue}{semantic agent} and synthesize motion via \textcolor{deepblue}{Human Motion Score Distillation (HMSD)} ($\mathcal{L}_{\text{HMSD}}$) from a pretrained motion diffusion model, producing natural text-aligned motion. The object is treated as a \textcolor{orange}{physical agent}, with its trajectory computed via \textcolor{orange}{MPM simulation}. At this stage, both agents move independently. \textbf{{(b) Physically-Aware Interaction Synthesis (Sec.~\ref{sec:interaction}):}} While continuing to optimize $\mathcal{L}_{\text{HMSD}}$ from (A), we coordinate the agents through a \textcolor{green}{Windowed Attraction Loss} ($\mathcal{L}_{\text{attr}}$) that guides the \textcolor{deepblue}{human} toward the \textcolor{orange}{object}. Upon \textcolor{green}{Contact} Detection, we trigger \textcolor{orange}{object re-simulation} with physically consistent momentum transfer. Finally, we render the composed \textcolor{green}{4D HOI} scene and apply \textcolor{green}{Video-SDS} ($\mathcal{L}_{\text{V-SDS}}$) to enhance \textcolor{green}{contact} fidelity.}
    \label{fig:overview}
     \vspace{-0.5cm}
\end{figure*}

\section{Related Work}
\label{sec:related}

\noindent \textbf{Text-to-4D Generation.} \quad Early text-to-4D methods primarily extended 2D diffusion priors to 3D representations via Score Distillation Sampling (SDS). 
DreamFusion \cite{poole2022dreamfusion} established the baseline using 2D priors, while subsequent approaches like DreamGaussian \cite{tang2023dreamgaussian} and GaussianDreamer \cite{yi2024gaussiandreamer} adopted 3D Gaussian Splatting (3DGS) for efficiency. To handle temporal consistency, 4D-fy \cite{bahmani20244d} and Consistent4D \cite{jiang2023consistent4d} introduced temporal attention, while recent work like CHORD \cite{lyu2026chord} extends these priors to multi-object choreography. Regardless, these methods rely purely on visual priors, resulting in inconsistent motions that ignore collisions. \\
\noindent \textbf{Human–Object Interaction (HOI) Generation.} 
OMOMO~\cite{li2023omomo} generates motion from object trajectories, paving the way for text-driven works like AvatarGO~\cite{cao2025avatargo} and InterDreamer~\cite{xu2024interdreamer}, which utilize contact retargeting and 2D priors. 
To improve synchrony, SyncDiff~\cite{he2024syncdiff} and HOIDiNi~\cite{ron2025hoidini} explicitly optimize geometric alignment. 
However, these purely kinematic methods lack physical modeling (mass, elasticity), treating objects as rigid props and failing to capture realistic deformations or prevent interpenetration. \\
\noindent \textbf{Generative 3D Animation.} \quad 
Recent approaches focus on animating static 3D assets. 
To this end, Animate3D~\cite{jiang2024animate3d} and AKD~\cite{li2025akd} utilize video diffusion models.
AnimateAnyMesh \cite{wu2025animateanymesh} performs feed-forward 3D asset animation while Animus3D~\cite{sun2025animus3d} introduces ``Motion Score Distillation''.

However, these methods operate on individual entities in non-physical environments. They fail to model the coupled physics of human-object interaction, frequently leading to scenes where contact is physically implausible or entirely absent. \\
\noindent \textbf{Physics-Based MPM \& Gaussian Splatting.} \quad Existing works \cite{xie2024physgaussian, huang2025dreamphysics, zhao2025physsplat} combine MPM with 3DGS to optimize physical properties but are restricted to single-object dynamics. In contrast, we apply this ``Neuro-Physical'' approach to a coupled system, utilizing simulation to enforce causal interaction between an articulated human and a deformable object.

\section{Method}
\label{sec:method}

Given a static 3D human and object represented as 3D Gaussian Splats (3DGS), along with a text prompt describing the desired human motion and a prompt describing the scene interaction, our goal is to synthesize a dynamic 4D scene where the human actively engages with the object in a physically plausible manner. As illustrated in Fig.~\ref{fig:overview}, our framework couples generative human motion with explicit physical simulation under a unified 3DGS representation (Sec.~\ref{sec:representation}). We synthesize motion independently for each agent (Sec.~\ref{sec:motion}), then coordinate them through attraction-based guidance, contact-driven re-simulation, and video prior distillation (Sec.~\ref{sec:interaction}). Implementation details are in the appendix and code will be made fully available.

\subsection{Scene Representation}
\label{sec:representation}

We adopt 3D Gaussian Splatting~\cite{kerbl20233d} as a shared representation for both agents, enabling joint rendering and optimization in a unified differentiable pipeline.

\noindent \textbf{3D Gaussian Splatting.}
3DGS represents scenes using a set of anisotropic Gaussians. Each Gaussian $\mathcal{G}_i$ is defined by position $\mathbf{x}_i$, covariance $\mathbf{\Sigma}_i$, opacity $\sigma_i$, and spherical harmonics $\mathbf{c}_i$ for view-dependent appearance. The color $\mathbf{C}$ of a pixel is computed by alpha-blending these 3D Gaussians when projected to the image plane: $\mathbf{C} = \sum_{i=1}^N T_i \alpha_i \mathbf{C}_i, \ \rm{with}\ T_i=\prod_{j=1}^{i-1} (1-\alpha_j)$, 

where $N$ is the set of depth-sorted Gaussian kernels affecting the pixel, and $C_i$ and $\alpha_i$ represents the color and density of this point computed by a 3D Gaussian $G$ with covariance $\mathbf{\Sigma}$ and opacity $\sigma$.

\noindent \textbf{Human as Semantic Agent.}
We represent the human using 3D Gaussians bound to the SMPL parametric body model~\cite{SMPL:2015}, following HUGS~\cite{kocabas2024hugs}. Each Gaussian is defined in an initial pose and deformed via Linear Blend Skinning (LBS). Given pose parameters $\boldsymbol{\theta}$ and joint transformations $\{\mathbf{G}_k\}_{k=1}^{K}$, the position $\boldsymbol{\mu}_i$ of Gaussian $i$ transforms as $\boldsymbol{\mu}'_i = \left( \sum_{k=1}^{K} w_{i,k} \mathbf{G}_k \right) \boldsymbol{\mu}_i$, 

where $w_{i,k}$ are skinning weights associating Gaussian $i$ with joint $k$, allowing direct optimization of pose parameters. 

\noindent \textbf{Object as Physical Agent.}
The object must respond to physical forces rather than learned priors.We treat its Gaussians as particles in a Material Point Method (MPM) simulation~\cite{stomakhin2013material, jiang2016material}, following PhysGaussian~\cite{xie2024physgaussian}, evolving positions $\mathbf{x}_i(t)$ according to continuum mechanics. Unlike the human, the object's motion is determined entirely by simulation, ensuring physical plausibility.

\subsection{Agent Motion Synthesis}
\label{sec:motion}

Having established the scene representation, we now synthesize motion for each agent, the physical agent via physical simulation, and the semantic agent via learned motion priors.

\noindent \textbf{Object Motion Simulation.}
The object's initial trajectory is computed via forward MPM simulation from $t=0$ to $T$, producing a physically consistent free-motion trajectory. This trajectory is updated once contact with the human is established (Sec.~\ref{sec:interaction}).

\noindent \textbf{Human Motion Score Distillation.}
We parameterize human motion as a sequence $X = \{x^t\}_{t=0}^{T}$, where each frame $x^t = (\mathbf{r}^t, \boldsymbol{\omega}^t, \boldsymbol{\theta}^t) \in \mathbb{R}^{D}$ consists of root translation $\mathbf{r}^t \in \mathbb{R}^3$, global orientation $\boldsymbol{\omega}^t \in \mathbb{R}^6$ in 6D rotation, and per-joint pose parameters $\boldsymbol{\theta}^t \in \mathbb{R}^{J \times 3}$ for $J$ joints. Given a pretrained Human Motion Diffusion Model (MDM)~\cite{tevet2022human} and a text prompt $p_{\text{motion}}$ describing the desired human motion, we define Human Motion Score Distillation (HMSD):
\begin{equation}
    \nabla_X \mathcal{L}_{\text{HMSD}} = \mathbb{E}_{t, \epsilon}\left[w_{HMSD}(t)\left(\hat{X}_0(X_t, t, p_{\text{motion}}) - X\right)\right],
\end{equation}
where $X_t$ is the motion $X$ corrupted with Gaussian noise $\epsilon \sim \mathcal{N}(0, I)$ at diffusion timestep $t$, $\hat{X}_0$ is the MDM's prediction of the clean motion conditioned on $X_t$ and the text prompt $p_{\text{motion}}$, and $w_{HMSD}(t)$ is a timestep-dependent weighting function. This objective pulls the optimized motion toward the manifold of natural human movements described by the text prompt. 
We optimize the human pose parameters using $\mathcal{L}_{\text{HMSD}}$ alone for $N_{\text{init}}$ iterations, producing natural human motion. However, at this stage, the motion is generated independently of the object's position and may not result in contact.

\subsection{Physically-Aware Interaction Synthesis}
\label{sec:interaction}

Given both agents' initial motions, the central challenge becomes coordinating them into a coherent interaction. We address this through three coupled mechanisms: (1) a windowed attraction loss for human-object coordination, (2) contact-driven re-simulation for physical response, and (3) distilling video priors for contact fidelity.

\noindent \textbf{Windowed Attraction Loss.}
To coordinate the generated motion with the object, we introduce a mechanism that identifies when and where contact should occur, then guides the relevant body part toward the object. This requires determining two quantities: the \textit{contact joint} $j^*$, i.e., which body part will make contact, and the \textit{contact frame} $t^*$, i.e., when impact should occur. We estimate both by analyzing the velocity profile of the initial motion. Intuitively, the joint most involved in the action exhibits the highest cumulative motion throughout the sequence, e.g. for a kick, this is the foot and for a punch, the hand. Contact should occur at the moment of peak velocity, as this is when the striking limb is maximally extended toward the target, transitioning from the acceleration phase to deceleration or follow-through. 

\begin{wrapfigure}{r}{0.5\textwidth}
    \vspace{-0.9cm} 
    \centering
    \includegraphics[width=0.48\textwidth]{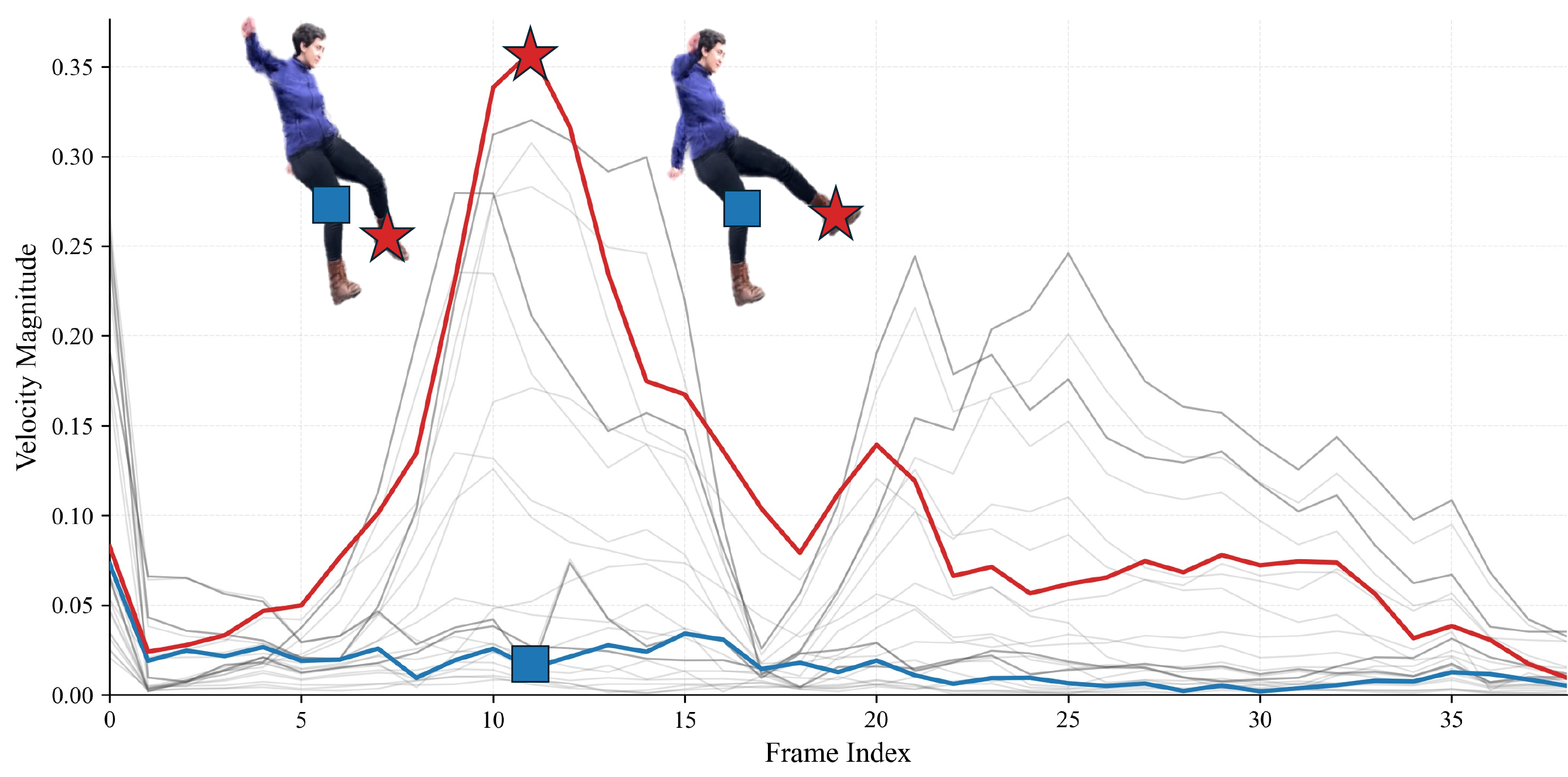}
    \caption{\textbf{Contact Joint and Frame Selection.} Per-joint velocity profiles for a kicking motion. Each curve represents a different SMPL joint, with the left foot (\textcolor{red}{$\bigstar$}) and right knee (\textcolor{royalblue}{$\blacksquare$}) highlighted. The left foot exhibits the highest cumulative velocity and is automatically selected as the contact joint $j^*$, with the contact frame $t^*$ identified at its peak. In contrast, the right knee (blue) maintains low velocity throughout the sequence, illustrating why it is not selected. We visualize human poses at two frames where peak motion occurs, illustrating the motion progression.}
    \label{fig:velocity_peak}
    \vspace{-0.5cm} 
\end{wrapfigure}
We demonstrate this intuition in Fig.~\ref{fig:velocity_peak}, where for a kicking motion, the foot joint exhibits both the highest cumulative velocity and a clear peak at the natural contact moment.

We first identify the contact joint by selecting the joint with highest cumulative velocity across all frames, then determine the contact frame as the moment of peak velocity for that joint:
\begin{equation}
    j^* = \operatorname*{argmax}_j \sum_{t=0}^{T-1} v_j(t), \quad t^* = \operatorname*{argmax}_t \, v_{j^*}(t),
\end{equation}
where $\mathbf{p}_j(t)$ is the world-space position of joint $j$ at frame $t$ obtained from SMPL forward kinematics, and $v_j(t) = \|\mathbf{p}_j(t+1) - \mathbf{p}_j(t)\|$ is its per-frame velocity. 
We then apply a Gaussian-weighted attraction loss that pulls the contact joint $j^*$ toward the object, with guidance concentrated around the contact frame $t^*$ while allowing natural motion elsewhere:
\begin{equation}
    \mathcal{L}_{\text{attr}} = \frac{\sum_{t} g(t) \|\mathbf{p}_{j^*}(t) - \mathbf{c}_{\text{obj}}(t)\|^2}{\sum_{t} g(t)}, \quad g(t) = \exp\left(-\frac{(t - t^*)^2}{2\sigma^2}\right),
\end{equation}
where $\mathbf{p}_{j^*}(t)$ is the position of the contact joint at frame $t$, $\mathbf{c}_{\text{obj}}(t)$ is the object's center of mass, and $g(t)$ is a Gaussian weighting function within a window, $[t^* - \Delta t^*, t^* + \Delta t^*]$, of the contact frame $t^*$,  with standard deviation $\sigma$. The Gaussian weighting concentrates guidance around the predicted contact moment while allowing the motion prior to govern the natural wind-up and follow-through phases without interference.

We continue optimization for $N_{\text{sync}}$ iterations with the objective $\mathcal{L}_{\text{human}} = \lambda_{\text{HMSD}} \mathcal{L}_{\text{HMSD}} + \lambda_{\text{attr}} \mathcal{L}_{\text{attr}}$.

This couples the motion prior with scene awareness, yielding coordinated human-object motion. We optimize the underlying SMPL parameters $\boldsymbol{\theta}$ throughout, not joint positions directly.

\noindent \textbf{Contact Detection and Re-simulation.}
While the attraction loss ensures the human motion is coordinated with the object, the object itself is not yet affected by this interaction and continues to follow its initial free-motion trajectory. To achieve physically plausible dynamics and contact, we detect the contact event and re-simulate the object's response to the applied force. After $N_{\text{sync}}$ iterations, we identify the contact frame and recompute the object trajectory accordingly, optimization then proceeds with this updated motion.

To detect contact, we first assign each human Gaussian $\mathcal{G}_i$ to its dominant joint based on skinning weights, where $\mathcal{G}_i$ is associated with joint $j$ if $j = \operatorname*{argmax}_k w_{i,k}$. For each joint $j$, we compute its axis-aligned bounding box $\mathcal{B}_j(t)$ from the positions of its associated Gaussians at frame $t$, and similarly compute the object's bounding box $\mathcal{B}_{\text{obj}}(t)$. We identify contact at frame $t_c$ with joint $j_c$ when: (1) $\mathcal{B}_{j_c}(t_c) \cap \mathcal{B}_{\text{obj}}(t_c) \neq \emptyset$, and (2) at least $\tau_{\text{contact}}$ fraction of joint $j_c$'s Gaussians lie within distance $d_{\text{contact}}$ of the nearest object Gaussian.

\begin{figure}[t]
    \centering
    \includegraphics[width=1\textwidth]{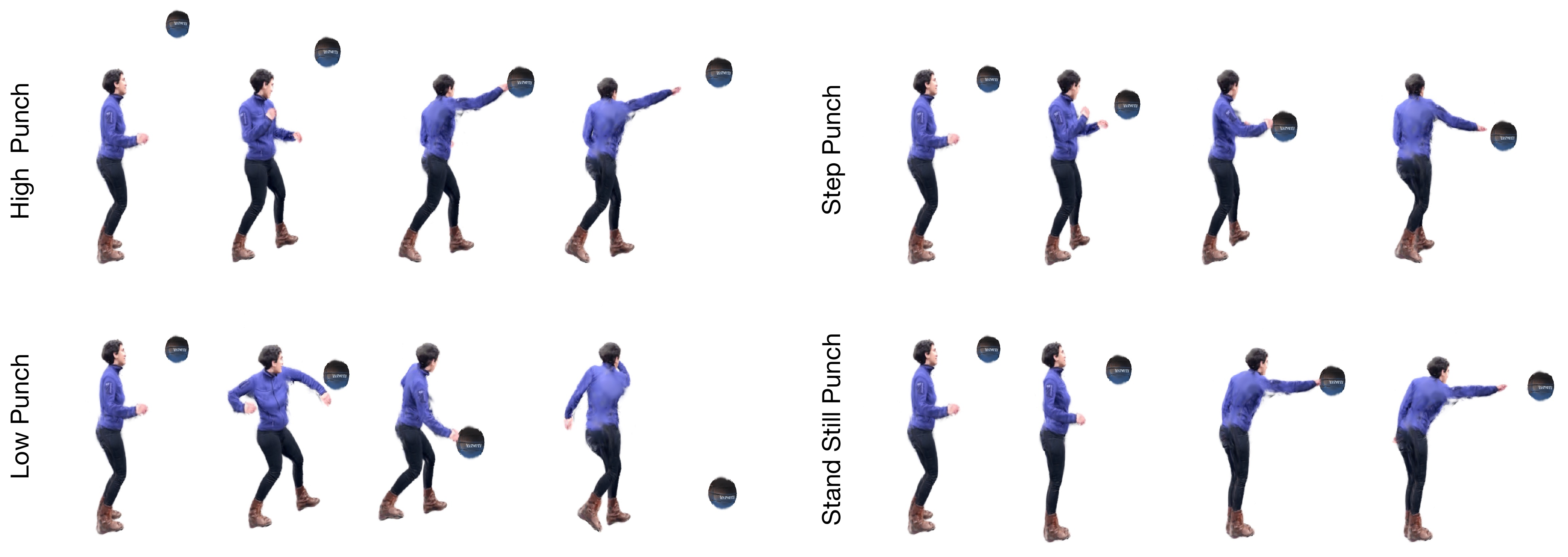} \\ 
    \caption{
    \textbf{In-Scene Variations}. We demonstrate controllability by varying human/object movements. \textit{Top \& Second Rows}: Changing object position (High vs. Low) forces trajectory adaptation. \textit{Third \& Bottom Rows}: Altering intensity (Step vs. Stand Still) yields distinct impact velocities. 
    }
    \vspace{-0.5cm}
\label{fig:variations}
\end{figure}

Once contact is detected, we compute the momentum transfer and update the object's velocity. We estimate the human velocity $\mathbf{V}_{\text{human}}$ from the contact joint's displacement. The contact normal $\mathbf{n}$ is defined as the direction from the mean position of contacting object Gaussians toward the object's center of mass. The post-impact velocity is then:
\begin{equation}
    v_{\text{in}} = (\mathbf{V}_{\text{human}} - \mathbf{V}_{\text{obj}}) \cdot \mathbf{n}, \quad
    \mathbf{V}_{\text{post}} = \mathbf{V}_{\text{obj}} + (1 + e) \, v_{\text{in}} \, \mathbf{n},
\end{equation}
where $e$ is the coefficient of restitution.
 We perform a single forward MPM simulation from $t_c$ to $T$ with the post-impact velocity, producing a physically consistent trajectory that respects momentum transfer and material properties. This simulated trajectory is then held fixed, such that subsequent optimization adjusts only human pose parameters, ensuring the object's response remains physically consistent. Additional details are provided in the appendix. 

\noindent \textbf{Video-SDS for Contact Fidelity.}
The contact region may still exhibit artifacts due to the discrete nature of contact detection and the independent optimization of human and object. Since both agents share a 3DGS representation, we can render the composed scene and apply Video Score Distillation Sampling~\cite{bahmani20244d} to enhance contact fidelity. Utilizing the $v$-prediction formulation from~\cite{li2025akd}, given rendered frames $V = \{I^t\}_{t=1}^{T}$ from sampled viewpoints, we encode them into latent space $z = \mathcal{E}(V)$, where $\mathcal{E}$ is the pretrained VAE encoder, and define the diffusion loss as:
\begin{equation}
    \mathcal{L}_{\text{Diff}}(z, p_{\text{scene}}) = \mathbb{E}_{t, \epsilon} \left[ w_{SDS}(t) \left\| z - \hat{z} \right\|_2^2 \right],
\end{equation}
where $\hat{z} = \sqrt{\alpha_t} z_t - v_{\phi}(z_t; t, p_{\text{scene}})$ is the reconstruction based on the predicted velocity $v_{\phi}$ from the pretrained video diffusion model, $p_{\text{scene}}$ is a text prompt describing the interaction, and $ w_{SDS}(t)$ is a timestep-dependent weighting function. Omitting the gradient through the velocity-predicting transformer, we optimize human pose parameters $\boldsymbol{\theta}$ via:
\begin{equation}
    \nabla_{\boldsymbol{\theta}} \mathcal{L}_{\text{V-SDS}} = \mathbb{E}_{t, \epsilon} \left[  w_{SDS}(t) \left( z - \hat{z} \right) \frac{\partial z}{\partial \boldsymbol{\theta}} \right].
\end{equation}
We apply temporal masking, optimizing only frames within a window $[t_c - \Delta t, t_c + \Delta t]$ around the contact frame, focusing optimization on contact frames while preserving the motion prior's influence elsewhere. Additional Video-SDS details are in the appendix.

\noindent \textbf{Optimization.} Our optimization proceeds in three stages: (1) $N_{\text{init}}$ iterations of $\mathcal{L}_{\text{HMSD}}$ to establish natural motion, (2) $N_{\text{sync}}$ iterations of $\mathcal{L}_{\text{human}} = \lambda_{\text{HMSD}}\mathcal{L}_{\text{HMSD}} + \lambda_{\text{attr}}\mathcal{L}_{\text{attr}}$ to coordinate with the object, followed by contact detection and MPM re-simulation, and (3) temporally-masked Video-SDS around contact frames to enhance contact fidelity. Additional details are in the supplementary.

\begin{figure*}[t!]
    \centering
    \includegraphics[width=1\textwidth]{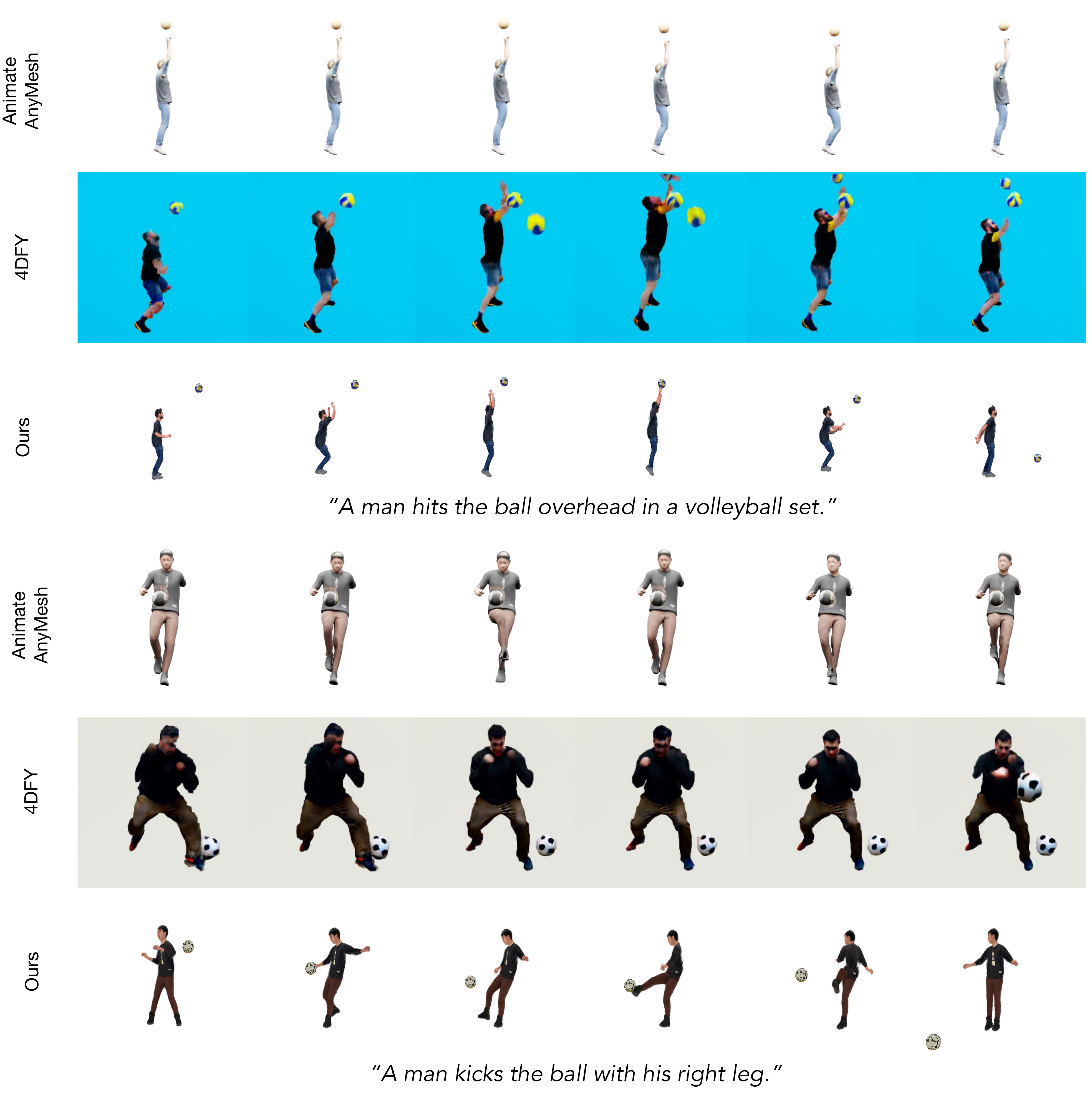} \\

    \caption{\textbf{Baseline Comparison}. We show a single view (see more views in appendix). While baselines exhibit missing contact (top) or ghosting artifacts (middle), our method (bottom) produces coherent interactions with causal momentum transfer and accurate physical response.}
    \vspace{-0.4cm}
\label{fig:comparisons}
\end{figure*}

\section{Experiments}
\label{sec:experiments}

We evaluate \method{} on diverse human-object interaction scenarios. We present qualitative results demonstrating the range of supported actions, humans, and objects in Sec.~\ref{sec:results}, compare against state-of-the-art baselines in Sec.~\ref{sec:comparisons}, and provide ablation studies in Sec.~\ref{sec:ablations}. We discuss limitations in the appendix.

\subsection{Interaction Generation}
\label{sec:results}

Fig.~\ref{fig:teaser} demonstrates our method's ability to generate physically plausible 4D human-object interactions across a variety of scenarios. We showcase multiple action types including \textit{punching}, \textit{kicking}, and \textit{pushing}, paired with different objects such as basketballs, soccer balls, file cabinets, etc. For each scenario, our framework successfully coordinates the human motion with the object trajectory, producing realistic interactions where the object responds according to its material properties. Across all examples, our method eliminates the ghosting and interpenetration artifacts common in purely generative approaches, while capturing dynamic object responses that kinematic methods cannot achieve. To further demonstrate controllability and physical consistency, we show in-scene variations in Fig.~\ref{fig:variations}, including different initial object velocities, positions, and contact intensities. These variations highlight that our framework produces coherent, physically plausible results across a range of initial conditions. Additional visualizations are provided in the supplementary material.

\subsection{Quantitative and Qualitative Evaluation}
\label{sec:comparisons}

We assemble a benchmark of 
10 distinct human-object interaction scenarios 
spanning different humans, objects, and interactions. For each combination, we generate 4D interactions and evaluate physical plausibility, semantic alignment, and visual quality.

\noindent \textbf{Baselines.}

We compare against 4D-fy \cite{bahmani20244d} and AnimateAnyMesh \cite{wu2025animateanymesh}, representing the most relevant baselines with available implementations. 4D-fy lacks explicit physics, leading to ghosting artifacts, while AnimateAnyMesh lacks coordination, frequently missing contact. We note that directly relevant HOI and 4D generation methods (AvatarGO~\cite{cao2025avatargo}, InterDreamer~\cite{xu2024interdreamer}, CHORD ~\cite{lyu2026chord}) lack publicly available code, so we compare against the strongest available methods spanning generative and animation paradigms.

\noindent \textbf{Metrics.}
We employ metrics that assess both semantic alignment and temporal quality of the generated interactions. \textit{ViCLIP}~\cite{wang2023internvid} measures semantic alignment between rendered videos and text prompts via cosine similarity in the joint video-text embedding space, providing a measure of how well the generated interaction matches the intended action.

\begin{wraptable}{r}{0.55\textwidth}
    \vspace{-0.2cm} 
    \centering
    \caption{\textbf{Quantitative Evaluation.} Comparison on VQA Phys., ViCLIP, and User Study MOS (Q1-Q4).}
    \label{tab:baselines}
    
    \resizebox{1.0\linewidth}{!}{ 
        \setlength{\tabcolsep}{1.5pt} 
        \begin{tabular}{lcccccc}
        \toprule
        Method & VQA & ViCLIP & Q1 & Q2 & Q3 & Q4 \\
        \midrule
        4DFY & 0.15 & 0.26 & 1.42 & 1.44 & 1.85 & 1.76\\
        AnimateAnyMesh & 0.19 & 0.24& 1.61 & 1.51 & 2.40 & 2.11 \\
        \midrule
        \textbf{Ours} & \textbf{0.25} & \textbf{0.30} & \textbf{4.33} & \textbf{4.29} & \textbf{4.21} & \textbf{4.04}\\
        \bottomrule
        \end{tabular}
    }
    \vspace{-0.3cm} 
\end{wraptable}
To evaluate physical realism, we employ a \textit{VQA Physics Score}~\cite{lin2024evaluating}, where using a VLM (Qwen-VL-7B), one queries: ``Is the interaction physically plausible overall?'' and reports the probability of the token ``Yes''.

In addition, we conduct a \textit{user study}, evaluating the perceptual quality of our method against baselines. Participants were presented with videos and asked to rate each method on a scale of 1 (worst) to 5 (best) based on four criteria: 
(Q1) \textit{Physical Plausibility} of the object's response to physics; 
(Q2) \textit{Contact Quality}, assessing the accuracy and realism of the interaction; 
(Q3) \textit{Motion Naturalness} of the human agent; and 
(Q4) \textit{Photorealism} of the visual appearance. 
We collected responses from 23 participants and report MOS scores.

\noindent \textbf{Qualitative Evaluation.}
A qualitative comparison is shown in Fig.~\ref{fig:comparisons}. 4D-fy struggles to maintain object consistency, often hallucinating multiple instances of the object throughout the sequence, while producing minimal human motion that fails to convey the intended action. AnimateAnyMesh generates limited motion for both human and object, with no meaningful contact occurring between them. In contrast, our method produces dynamic human motion that coordinates with the object, achieving proper contact where the object responds with physically plausible trajectories and material-appropriate dynamics. 

\noindent \textbf{Quantitative Evaluation.}
Tab.~\ref{tab:baselines} presents the quantitative comparisons to baselines. 
Our method achieves the highest scores on all metrics, significantly outperforming baselines on VQA Physics ($0.253$ vs. $0.196$), ViCLIP ($0.295$ vs. $0.256$) and in perceptual studies. 

\begin{figure*}[t!]
    \centering
    \includegraphics[width=0.95\textwidth]{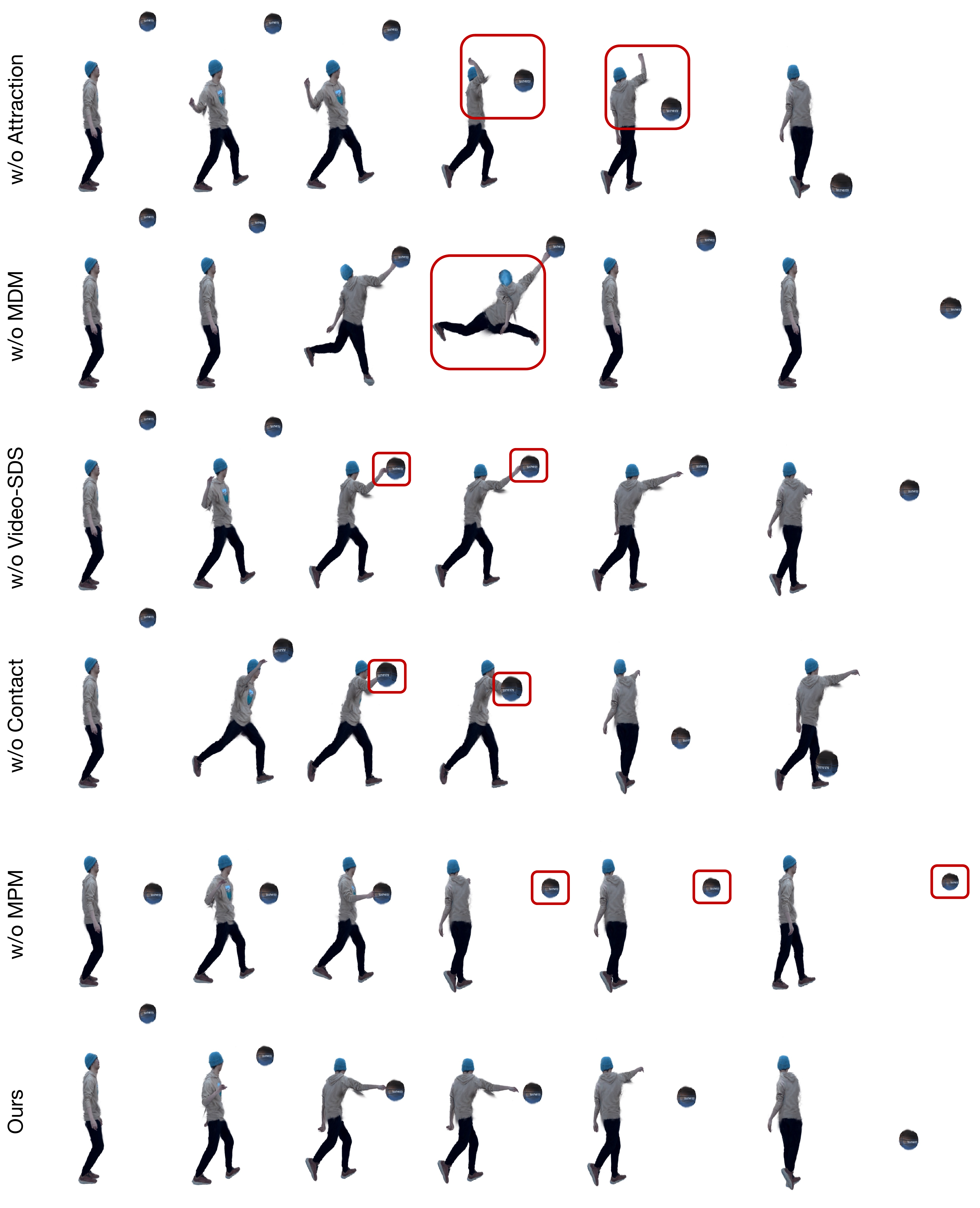}

\caption{\textbf{Qualitative Ablation}. 
We highlight failure cases when removing components of our method (see highlighted \textit{\textcolor{red}{boxes}} emphasizing the failure). 
\textit{w/o Attraction:} The agent fails to hit the object. 
\textit{w/o MDM:} The human mesh deforms unnaturally. 
\textit{w/o Video-SDS:} Severe penetration occurs. 
\textit{w/o Contact:} The hand passes through the object. 
\textit{w/o MPM:} The object moves via velocity transfer, lacking physical realism.} 
\label{fig:ablations}
\vspace{-0.3cm}
\end{figure*}

\subsection{Ablation Study}
\label{sec:ablations}

We validate the necessity of individual components of our method in Tab.~\ref{tab:ablations}, considering both automated metrics and a perceptual user study as noted above. We also visualize their effect for a single example in Fig.~\ref{fig:ablations}. 
Removing Video-SDS (\textit{w/o Video-SDS}) preserves global physics but leaves local penetration artifacts due to discrete contact detection.

\begin{wraptable}{r}{0.55\textwidth}
    \vspace{-0.6cm} 
    \centering
    \caption{\textbf{Ablation Study.} Impact of components on automated metrics and User Study MOS scores (Q1--Q4).}
    \label{tab:ablations}
    
    \resizebox{1.0\linewidth}{!}{
        \setlength{\tabcolsep}{2.5pt} 
        \begin{tabular}{lcccccc}
        \toprule
        Method & VQA & ViCLIP & Q1 & Q2 & Q3 & Q4 \\
        \midrule
        w/o Attraction & 0.24 & 0.23 & 2.46 & 1.85 & 2.97 & 2.83\\
        w/o Contact & 0.20 & 0.23 & 1.92 & 1.61 & 2.83 & 2.55\\
        w/o MDM & 0.20 & 0.22 & 3.26 & 2.82 & 1.85 & 2.65 \\
        w/o Video-SDS & 0.25 & 0.27 & 4.13 & 4.15 & 4.11 & 3.75\\
        w/o MPM & 0.21 & 0.23 & 3.04 & 3.58 & 3.65 & 3.56\\
        \midrule
        \textbf{Ours (Full)} & \textbf{0.25} & \textbf{0.30} & \textbf{4.41} & \textbf{4.44} & \textbf{4.25} & \textbf{4.15}\\
        \bottomrule
        \end{tabular}
    }
    \vspace{-0.4cm} 
\end{wraptable}

Removing the windowed attraction loss (\textit{w/o Attraction}) decouples the agent from the scene, causing the otherwise natural human motion to miss the target entirely. 
Replacing the motion diffusion prior with direct pose optimization (\textit{w/o MDM}) produces unnatural, anatomically implausible motion to satisfy attraction constraints. 
Disabling contact detection and re-simulation 
(\textit{w/o Contact}) breaks causality; the human reaches the object, but the object ignores the collision and continues its trajectory.
Finally, removing MPM simulation entirely (\textit{w/o MPM}) reduces object dynamics to constant velocity, losing material-aware physical fidelity.
Notably, the metrics reflect the ablations' failure modes. The \textit{w/o Attraction} variant retains high VQA Physics as motion remains natural, but drops in ViCLIP due to absent contact; \textit{w/o Contact} scores lowest on physics as ignored collisions are the most salient violation, and \textit{w/o MDM} suffers in prompt alignment since only frames within the optimization window exhibit motion.

\section{Conclusion}

We presented \method, a framework that couples generative human motion with MPM-based physical simulation under a shared 3DGS representation to produce physically plausible 4D human-object interactions. Experiments show that this neuro-physical coupling eliminates ghosting and interpenetration artifacts while enabling dynamic post-contact object responses, outperforming existing baselines in text alignment, physical plausibility, and contact quality. We believe bridging data-driven generation with physics-based simulation opens promising avenues for realistic 4D content creation.

\clearpage

\bibliographystyle{unsrtnat} 
\bibliography{references}

\newpage
\appendix

\section{Interactive Visualizations}
We refer readers to the interactive visualizations on our project page at \url{https://omerbenishu.github.io/PhyGenHOI/} for full temporal sequences of generated 4D human-object interactions, comparisons with baselines, and ablation studies across diverse action types.
\section{Additional Details}
\label{sec:supp_details}

\subsection{Implementation Details}
\label{sec:imp_details}

\medskip \noindent \textbf{Hardware and Runtime.}
All experiments are conducted on a single NVIDIA H200 GPU. The full pipeline runtime per scene is approximately 74 minutes, broken down as follows: human motion optimization takes about 10 minutes, MPM simulation takes 4 minutes, and Video-SDS refinement takes approximately 1 hour. Rendering the final 4D sequence achieves 20 FPS.

\medskip \noindent \textbf{Human Representation.}
We represent the human using 3D Gaussians bound to the SMPL parametric body model~\cite{SMPL:2015}, following HUGS~\cite{kocabas2024hugs}. We use the default hyperparameters provided by the authors \url{https://github.com/apple/ml-hugs}. Each human is initialized from a pre-trained HUGS model, in which the Gaussian representation is already learned and coupled with the SMPL parameters.

\medskip \noindent \textbf{Object Representation.}
Object 3DGS representations are obtained from from two sources. The blue ball object is taken directly from the DreamPhysics \cite{huang2025dreamphysics} dataset hosted on Hugging Face, while all other objects are reconstructed from single images using Trellis \cite{xiang2024structured} image-to-3D pipeline. We use the standard 3D Gaussian Splatting~\cite{kerbl20233d} representation with default parameters from the code provided by the authors \url{https://github.com/graphdeco-inria/gaussian-splatting}.

\subsection{Human Motion Score Distillation Details}
\label{sec:hmsd_details}

We employ the Motion Diffusion Model (MDM)~\cite{tevet2022human} as our human motion prior, and we use the pretrained model from STMC~\cite{petrovich24stmc} (\url{https://github.com/nv-tlabs/stmc}), which operates directly in SMPL pose space, enabling seamless integration with our 3DGS human representation.

\medskip \noindent \textbf{Motion Representation.}
Human motion is parameterized as a sequence $X = \{x^t\}_{t=0}^{T}$, where each frame $x^t = (\mathbf{r}^t, \boldsymbol{\omega}^t, \boldsymbol{\theta}^t)$ consists of root translation $\mathbf{r}^t \in \mathbb{R}^3$, global orientation $\boldsymbol{\omega}^t \in \mathbb{R}^6$ in 6D rotation representation, and per-joint pose parameters $\boldsymbol{\theta}^t \in \mathbb{R}^{J \times 3}$ for $J = 24$ joints. We generate sequences of $T = 40$ frames at $20$ FPS.

\medskip \noindent \textbf{Score Distillation.}
For HMSD, we sample diffusion timesteps uniformly from $[t_{\min}, t_{\max}]$, where $t_{\min} = 0$ and $t_{\max} = 100$. The weighting function is defined as $w(t) = 1 - \bar{\alpha}_t$. We use classifier-free guidance with a scale of $7.5$.

\subsection{Contact Detection and Re-simulation Details}
\label{sec:contact_details}

\medskip \noindent \textbf{Contact Detection.}
As described in Sec. 3.3 of the main paper, we detect contact by first assigning each human Gaussian to its dominant joint based on skinning weights. Contact at frame $t_c$ with joint $j_c$ is identified when two conditions are satisfied:
\begin{enumerate}
    \item The axis-aligned bounding boxes overlap: $\mathcal{B}_{j_c}(t_c) \cap \mathcal{B}_{\text{obj}}(t_c) \neq \emptyset$
    \item At least $\tau_{\text{contact}} = 0.05$ fraction of joint $j_c$'s Gaussians lie within distance $d_{\text{contact}} = 0.01$ of the nearest object Gaussian.
\end{enumerate}

\medskip \noindent \textbf{Velocity Update.}
Upon detecting contact, we compute the momentum transfer as follows. The human velocity $\mathbf{V}_{\text{human}}$ is estimated from the contact joint's displacement:
\begin{equation}
    \mathbf{V}_{\text{human}} = \frac{\mathbf{p}_{j_c}(t_c) - \mathbf{p}_{j_c}(t_c - 1)}{\Delta t},
\end{equation}
where $\Delta t = 1$. The contact normal $\mathbf{n}$ is computed as the normalized direction from the mean position of contacting object Gaussians toward the object's center of mass. The post-impact velocity applied to the object is:
\begin{equation}
    \mathbf{V}_{\text{post}} = \mathbf{V}_{\text{obj}} + (1 + e) \cdot v_{\text{in}} \cdot \mathbf{n},
\end{equation}
where $v_{\text{in}} = (\mathbf{V}_{\text{human}} - \mathbf{V}_{\text{obj}}) \cdot \mathbf{n}$ is the relative velocity along the contact normal, and $e$ is the coefficient of restitution.

\medskip \noindent \textbf{MPM Simulation Parameters.}
We base our MPM simulation on PhysGaussian~\cite{xie2024physgaussian}, using the Taichi~\cite{hu2019taichi} framework. We use a grid resolution of $64$ with simulation timestep $4\cdot10^{-5}$ and run for 1250 total steps per frame. Material properties are set with Young's modulus $10^{7}$ and Poisson ratio $0.45$. The coefficient of restitution $e$ is $0.6$. After contact detection at frame $t_c$, we perform a single forward MPM simulation from $t_c$ to $T$ with the computed post-impact velocity, producing the final object trajectory.

\subsection{Video-SDS Details}
\label{sec:vsds_details}

We employ Video Score Distillation Sampling using CogVideoX-5B \cite{yang2024cogvideox} as the video diffusion model. 

\medskip \noindent \textbf{Rendering and Sampling.}
During optimization, we render video clips at $480x720$ resolution with a length of $49$ frames from randomly sampled training camera viewpoints. Camera viewpoints are sampled uniformly from a circular trajectory around the scene, maintaining a fixed elevation for all cameras. A total of $100$ viewpoints are used, evenly spaced along the circle to ensure uniform coverage.

\medskip \noindent \textbf{Temporal Masking.}
We apply temporal masking to focus optimization on contact frames while preserving the motion prior's influence elsewhere. Specifically, we optimize only frames within a window $[t_c - \Delta t, t_c + \Delta t]$ around the contact frame $t_c$, where $\Delta t = 1$  frames.

\medskip \noindent \textbf{Diffusion Parameters.}
We apply a classifier-free guidance (CFG) scale of $100$. For timestep sampling, we sample uniformly from $[t_{\text{min}}, t_{\text{max}}]$ where $t_{\text{min}} = 100$ and $t_{\text{max}} = 980$, and the maximum timestep decreases linearly, reaching $t_{\text{max}} = 300$ at iteration $1000$. The text prompt $\mathbf{p}_{\text{scene}}$ describes the interaction. The prompt is rich, serving as the standard for the video models, with particular attention in the negative prompt to ensure realistic contact and avoid any penetration.

\subsection{Optimization Details}
\label{sec:opt_details}

Our optimization proceeds in three stages as described in Sec. 3.3 of the main paper.

\medskip \noindent \textbf{Stage 1: Motion Initialization.}
We optimize human pose parameters using $\mathcal{L}_{\text{HMSD}}$ alone for $N_{\text{init}} = 100$ iterations. We use the Adam optimizer with learning rate $0.005$.

\medskip \noindent \textbf{Stage 2: Human-Object Coordination.}
We continue optimization for $N_{\text{sync}} = 200$ iterations with the combined objective:
\begin{equation}
    \mathcal{L}_{\text{human}} = \lambda_{\text{HMSD}}\mathcal{L}_{\text{HMSD}} + \lambda_{\text{attr}} \mathcal{L}_{\text{attr}},
\end{equation}
where $\lambda_{\text{HMSD}} = 10.0$ and $\lambda_{\text{attr}} = 1.0$. The Gaussian window standard deviation for the attraction loss is set to $\sigma = 2$ frames. Learning rate remains $0.005$.

Following this stage, we perform contact detection and MPM re-simulation as described in Sec.~\ref{sec:contact_details}. The object trajectory is then fixed for subsequent optimization.

\medskip \noindent \textbf{Stage 3: Video-SDS for Contact Fidelity.}
We optimize human pose parameters using temporally-masked Video-SDS for $3000$ iterations with learning rate $0.001$. Video diffusion model details, rendering parameters, temporal masking window, and CFG scale are provided in Sec.~\ref{sec:vsds_details}.

\subsection{Comparisons and Ablations}
\label{sec:comp_details}

Below we provide details needed to reproduce the comparisons and ablations shown in the paper.

\medskip \noindent \textbf{4D-fy}~\cite{bahmani20244d}
We use the code provided by the authors \url{https://github.com/sherwinbahmani/4dfy}. We follow the original configurations used in the paper, while additionally applying the authors' recommendation to increase motion by setting $\text{system.loss.lambda\_sds\_video = 0.5}$.

\medskip \noindent \textbf{AnimateAnyMesh}~\cite{wu2025animateanymesh}
We use the code provided by the authors \url{https://github.com/JarrentWu1031/AnimateAnyMesh}. Since the method takes meshes as input and we have 3DGS objects, we first rendered the scenes using \url{https://meshy.ai/} and exported them to GLB format. However, due to one of their limitations, the mesh face counts are relatively low. In addition, to simplify the setup and improve motion, we positioned the objects closer than in the original scenes, as objects placed further away resulted in barely any scene movement.

\medskip \noindent \textbf{Ablation Variants.}
We evaluate five ablation variants. \textbf{w/o Video-SDS} skips optimization stage 3 entirely (Video-SDS), using the output from optimization stage 2 directly. \textbf{w/o Attraction} sets $\lambda_{\text{attr}} = 0$ during optimizationtage 2, optimizing only with $\mathcal{L}_{\text{HMSD}}$. \textbf{w/o MDM} replaces MDM-based optimization with direct pose parameter optimization using only $\mathcal{L}_{\text{attr}}$ and Video-SDS, initialized from the given initial position. \textbf{w/o Contact} skips contact detection and MPM re-simulation, allowing the object to follow its initial free-motion trajectory throughout. \textbf{w/o MPM} replaces MPM simulation with constant-velocity linear trajectories for the object.

\begin{figure*}[t!]
    \centering
    \includegraphics[width=\textwidth]{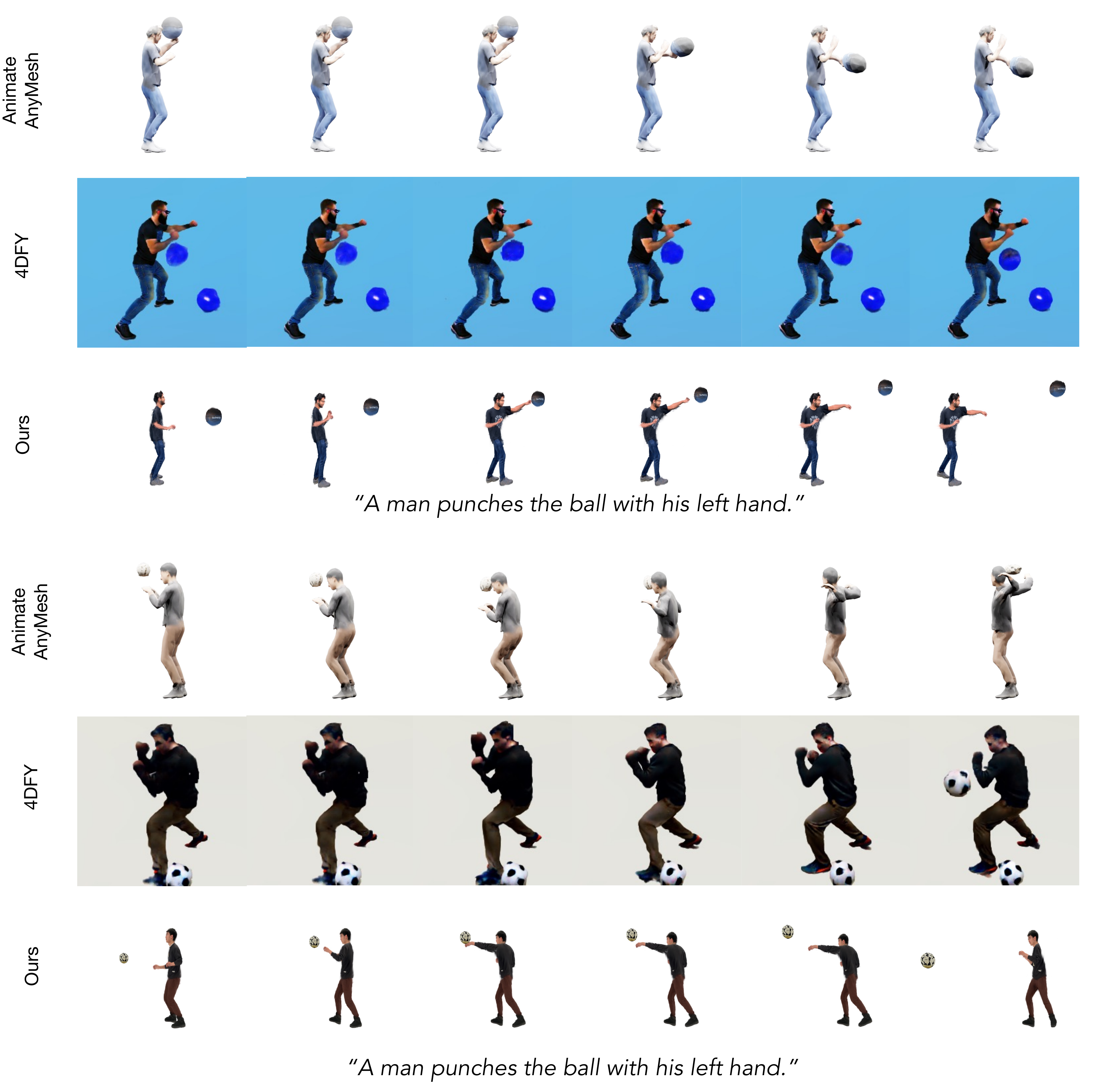}
    \caption{\textbf{Additional Comparisons.} Extended evaluation across diverse actions. Our framework consistently maintains physical causality and contact fidelity, whereas baselines fail to coordinate the human agent with the dynamic object. }
\label{fig:additional_comparisons}
\vspace{-0.5cm}
\end{figure*}
\clearpage

\section{Additional Qualitative Results}
\label{sec:additional_results}

Detailed visual comparisons across our full benchmark are provided in Fig.~\ref{fig:additional_comparisons}. As observed, our method successfully coordinates the human agent with the dynamic object across diverse action types, whereas baselines often struggle with physical causality or contact fidelity.

\section{Limitations}
\label{sec:limitations}

Our framework is designed for impulsive interactions such as kicking, punching, and pushing where contact triggers a discrete momentum transfer.
We note that continuous contact scenarios fall outside the current scope, as these require sustained force modeling rather than instantaneous impact.
Extending our formulation to handle such interactions is an interesting direction for future work. Our underlying formulation, however, is general, the windowed attraction and re-simulation components can naturally extend to multiple sequential contacts or multi-object scenes. Second, our attraction loss targets the object's center of mass, which is effective for convex objects, but may be suboptimal for complex geometries requiring contact at specific surface regions. Finally, while the object exhibits physical deformation via MPM, the human agent remains kinematic (SMPL), and thus does not respond to reaction forces or secondary collisions. Coupling soft-body simulation and two-way physical feedback to model bidirectional tissue deformation is a promising avenue for future work.

\section{Broader Impacts}
\label{sec:broader_impacts}

Our work on PhyGenHOI focuses on foundational research in physical simulation and 3D/4D generative methodologies. By enabling the synthesis of dynamic human-object interactions that are both visually faithful and physically plausible, our framework offers significant positive societal impacts for applications in animation, gaming, and immersive virtual reality. However, as with many advancements in generative modeling, we must acknowledge potential negative societal impacts. 

Specifically, improving the physical realism and causal accuracy of human actions, such as realistically depicting a person punching or kicking an object, could theoretically be misused by bad actors to generate highly convincing deepfakes for disinformation or malicious narratives. While our current method is not tied to a specific real-world deployment, mitigating these potential risks in future downstream applications will be important. Possible mitigation strategies include the gated release of interaction models, the integration of robust forensic watermarking on generated 4D assets, and the parallel development of advanced deepfake detection mechanisms to monitor potential misuse.

\end{document}